\documentclass[journal]{IEEEtran}

\usepackage{bbding}
\usepackage{makecell}
\usepackage{amsmath,amssymb}
\usepackage{graphicx}
\usepackage{setspace}
\usepackage{longtable}
\usepackage{colortbl,booktabs}
\usepackage{tabularx,multirow}
\usepackage{xcolor}
\usepackage{tikz}
\usepackage{multirow}
\usepackage{comment}
\graphicspath{{./pic/}}
\usepackage{color}
\usepackage{algorithm}
\usepackage{algorithmic}
 
\newlength\myindent
\usepackage{mathrsfs} 

\usepackage{array}  
\usepackage{booktabs}

\usepackage{hyperref}
\usetikzlibrary{fadings}
\usetikzlibrary{patterns}
\usetikzlibrary{shadows.blur}
\usetikzlibrary{arrows,shapes,chains}
\usepackage[numbers]{natbib}
\newcommand{\figref}[1]{Fig.~{\ref{#1}}}

\newcommand{\tableref}[1]{Table~{\ref{#1}}}
\newcommand{\sectionref}[1]{Section~{\ref{#1}}}

\makeatletter
\def\hlinewd#1{%
	\noalign{\ifnum0=`}\fi\hrule \@height #1 \futurelet
	\reserved@a\@xhline}
\makeatother

\begin{document}

\title{Transformation-Invariant Network for Few-Shot Object Detection in Remote Sensing Images}

\IEEEtitleabstractindextext{
\begin{abstract}

Object detection in remote sensing images relies on a large amount of labeled data for training. However, the increasing number of new categories and class imbalance make exhaustive annotation impractical. Few-shot object detection (FSOD) addresses this issue by leveraging meta-learning on seen base classes and fine-tuning on novel classes with limited labeled samples. Nonetheless, the substantial scale and orientation variations of objects in remote sensing images pose significant challenges to existing few-shot object detection methods. To overcome these challenges, we propose integrating a feature pyramid network and utilizing prototype features to enhance query features, thereby improving existing FSOD methods. We refer to this modified FSOD approach as a Strong Baseline, which has demonstrated significant performance improvements compared to the original baselines. Furthermore, we tackle the issue of spatial misalignment caused by orientation variations between the query and support images by introducing a Transformation-Invariant Network (TINet). TINet ensures geometric invariance and explicitly aligns the features of the query and support branches, resulting in additional performance gains while maintaining the same inference speed as the Strong Baseline. Extensive experiments on three widely used remote sensing object detection datasets, i.e., NWPU VHR-10.v2, DIOR, and HRRSD demonstrated the effectiveness of the proposed method. 
 
\end{abstract}

}

\markboth{Journal of \LaTeX\ Class Files,~Vol.~14, No.~8, August~2015}%
{Shell \MakeLowercase{\textit{et al.}}: Bare Demo of IEEEtran.cls for IEEE Journals}
	\author{Nanqing~Liu, Xun~Xu, Turgay~Celik, Zongxin~Gan, Heng-Chao Li
	\IEEEcompsocitemizethanks{
		\IEEEcompsocthanksitem 

        This work was supported in part by the National Natural Science Foundation of China under
        Grant 62271418, in part by the Natural Science Foundation of Sichuan Province under Grant 23NSFSC0030. This work was partially done during Nanqing~Liu's attachment with I2R, A*STAR. (Corresponding author: Xun Xu, Heng-Chao Li)

		Nanqing~Liu (lansing163@163.com),  Zongxin~Gan (gccxeon@my.swjtu.edu.cn), Heng-Chao Li (lihengchao\_78@163.com) are with the School of Information Science and Technology, Southwest Jiaotong University, Chengdu, China. 

  		Xun Xu (xux@i2r.astar.edu.sg) is with I2R, A-STAR, Singapore 138632, and the School of Information Science and Technology, Southwest Jiaotong University, Chengdu, China.
    
        Turgay Celik (celikturgay@gmail.com) is with the School of Information Science and Technology, Southwest Jiaotong University, Chengdu, China, and the School of Electrical and Information Engineering, University of the Witwatersrand, Johannesburg, South Africa.

}
	\thanks{Manuscript revised \today.}}
\maketitle
\IEEEdisplaynontitleabstractindextext

\section{Introduction}

\begin{figure*}[!htb]
	\centering
 
        \setlength\tabcolsep{3pt}
 	\begin{tabular}{ccccc}
  
		\includegraphics[width=0.315\linewidth]{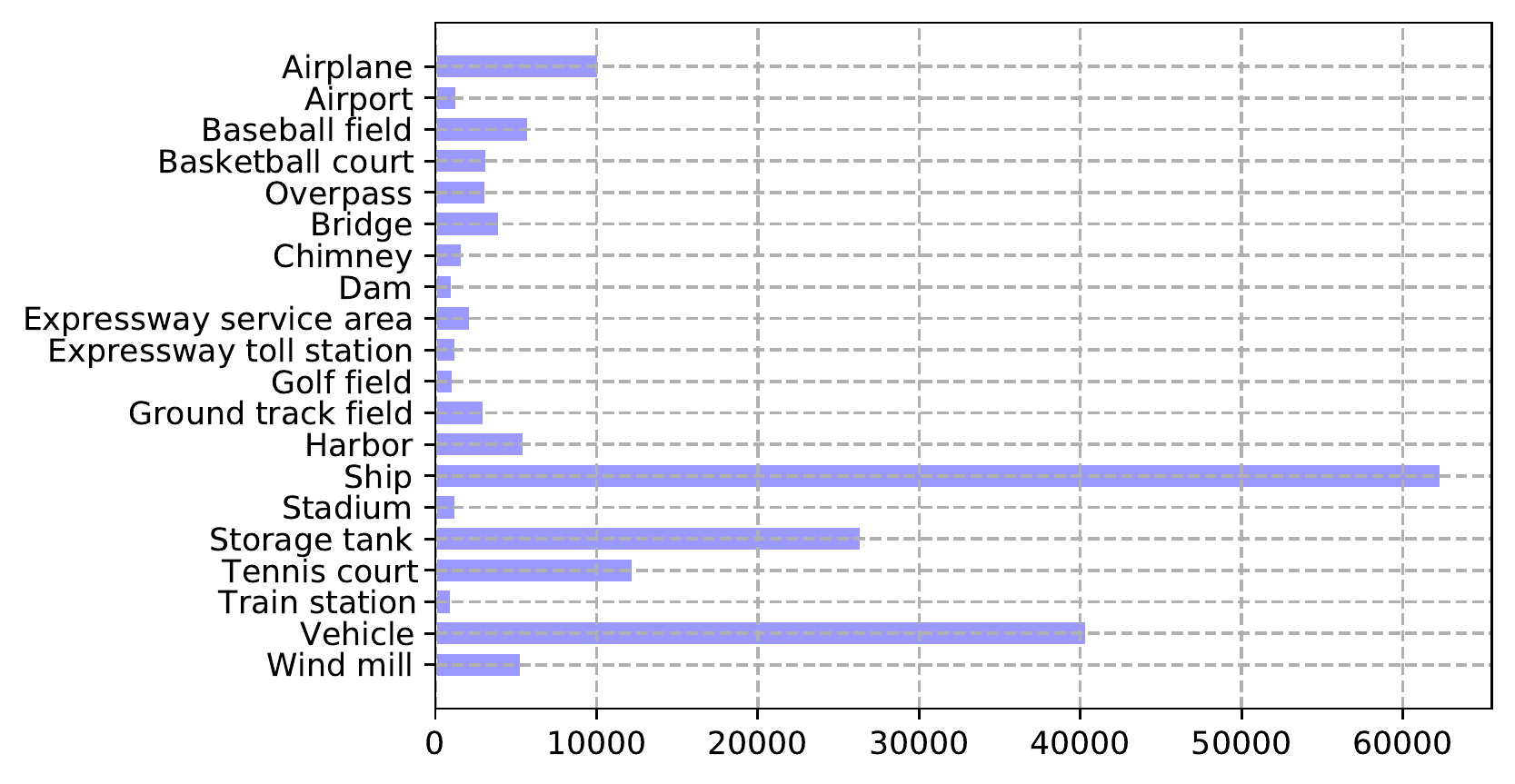}&
		\includegraphics[width=0.157\linewidth]{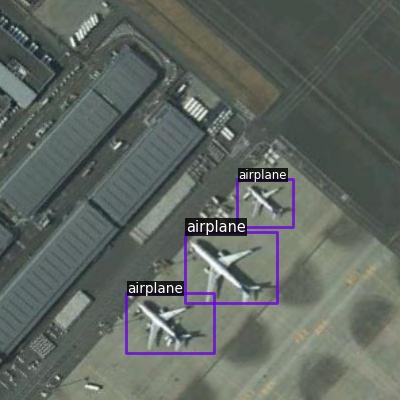}&
		\includegraphics[width=0.157\linewidth]{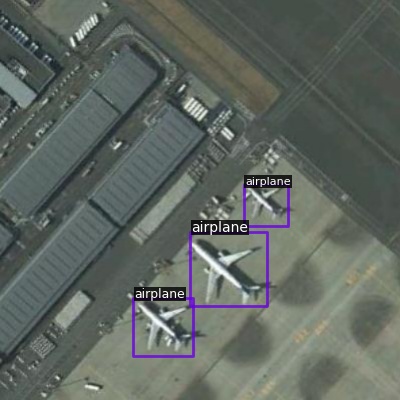}&
		\includegraphics[width=0.157\linewidth]{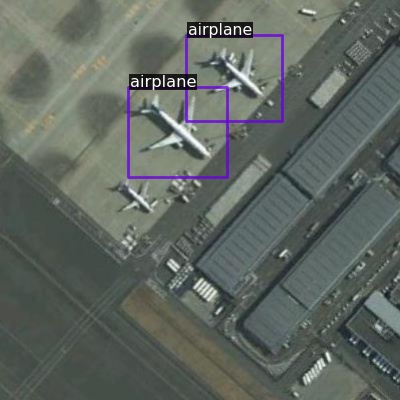}&
		\includegraphics[width=0.157\linewidth]{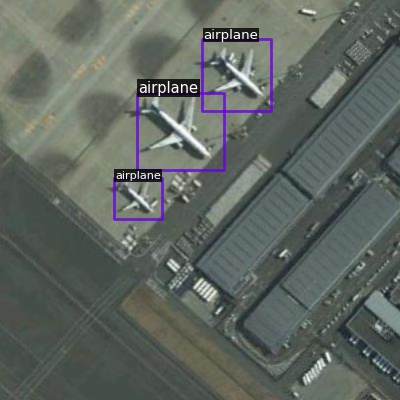} \\
        (a)&(b)&(c)&(d)&(e)
	\end{tabular}

	\caption{(a) Number of object instances per class in the DIOR dataset. (b-e) Comparison of the detection results of the Strong Baseline and TINet. (b) Detection results of the Strong Baseline on the original image. (c) Detection results of TINet on the original image. (d) Detection results of the Strong Baseline on the diagonally flipped image. (e) Detection results of TINet on the diagonally flipped image.}
	\label{fig:vis}

\end{figure*}

Optical remote sensing analyzes images captured by satellites and aerial vehicles. There are huge values for analyzing these remote sensing images~(RSIs), e.g. environmental monitoring \cite{chen2023continuous}, resource survey \cite{liu2020msnet}, and building extraction \cite{chen2021building}. Detecting natural and man-made objects from RSIs is the most critical capability that supports the above analytic tasks. The state-of-the-art approaches toward object detection in RSIs~\cite{qpdet, Liu2021, Yang2021, Cheng2021, glfpn, Han2022} employ a deep learning-based paradigm that requires a substantial amount of labeled data for supervised training. Nevertheless, there are several key challenges that prohibit standard supervised object detection training from scaling up. First, the existing object detection approaches \cite{10028728,lin2017,ren2015} often detect the objects from the seen semantic categories while the potential objects of interest are always non-exhaustive. When new categories of objects emerge, collecting enough labeled training data for novel categories is prohibitively expensive. 
Moreover, there are many classes with low quantities in RSIs, as evidenced by the frequency of objects in the DIOR dataset~\cite{li2020} in Fig.~\ref{fig:vis}~(a). This observation suggests that even if exhaustive annotation might be possible, collecting enough training examples for the minority classes is non-trivial which again motivates us to explore learning from a few labeled examples. This can help reduce the demand for annotated data and better adapt to the detection tasks of unknown and low-frequency categories. Common techniques used for limited annotation and unknown class learning include few-shot learning \cite{cheng2021spnet,lang2023base,cheng2022holistic}, zero-shot learning \cite{zsod, xu2017transductive, rsprompter}, and open vocabulary learning \cite{chen2023ovarnet,zareian2021open} and so on.

In this paper, we adopt the Few-Shot Object Detection (FSOD) paradigm to address the aforementioned challenges. In the field of object detection in natural semantic images, meta-learning-based FSOD methods \cite{yan2019, fsdetview, kang2019,vfa} have been extensively studied, which mainly consist of two branches: the query branch and the support branch. The query branch learns the object detection task from query images, while the support branch provides auxiliary information and feature representation for base and novel categories, allowing the query branch to better adapt to variations in different object categories. The interaction learning mechanism between the query and support features enables the few-shot object detection methods to have stronger generalization adaptability. However, directly applying existing meta-learning-based FSOD methods to RSIs is not the optimal choice for two reasons. Firstly, objects in images exhibit significant scale variations, and fixed-resolution object detection cannot effectively generalize to objects with large-scale variations. Secondly, the orientation variations in RSIs are more diverse than objects in natural images, as the cameras are positioned at a vertical angle, allowing arbitrary rotations in the XOY plane. This will cause spatial misalignment between the query and support images. Since the query branch learns from query images, which may have different orientations, it is difficult for it to effectively aggregate the feature representation provided by the support branch.

To address these challenges, we have made improvements to existing meta-learning-based FSOD methods. Previous meta-learning methods \cite{yan2019, fsdetview} only utilize C4-layer to generate proposals from the backbone network. Therefore, to adapt to scale variations in RSIs, we introduce the Feature Pyramid Network (FPN) \cite{lin2017} into the existing meta-learning methods. We further propose to highlight query feature maps with support prototype features through depth-wise convolution. These methods are simple but effective and significantly improve the performance of FSOD on RSIs. We then refer to the modified FSOD method as the Strong Baseline.

In addition to the Strong Baseline, we further address the challenges posed by large orientation variations by introducing a Transformation-Invariant Network (TINet). Specifically, we observe that the Strong Baseline cannot adapt well to object orientation. Therefore, we utilize both the query image and its transformed version as inputs to the network. Then a one-to-one consistency constraint is used to supervise the predicted bounding boxes of the original and transformed images. With these operations, the TINet is forced to produce consistent predictions on input images agnostic to camera poses. This explicitly aligns the spatial features of the query branch and the support branch. As a result, TINet can better identify objects with more pose variations. For example, in Fig. \ref{fig:vis} (b)(c)(d)(e), we demonstrate the prediction results of the Strong Baseline and TINet on different inputs. It can be observed that TINet can adapt well to perturbations caused by transformed images. However, the Strong Baseline fails to locate all airplanes accurately.

To evaluate our proposed methods, we conduct extensive experiments on the  DIOR \cite{li2020}, HRRSD \cite{hrrsd} and NWPU VHR-10.v2 \cite{li2017} datasets. The proposed TINet achieved the state-of-the-art few-shot object detection performance on all the above datasets. The main contributions of this paper are summarized as follows:

\begin{itemize}

\item Motivated by the large-scale variation, we proposed a Strong Baseline few-shot object detection method, which incorporates an FPN and uses 1$\times$1 depth-wise convolution to aggregate query and support features. With these operations, the Strong Baseline improves significantly from previous meta-learning FSOD approaches. 

\item We propose a transformation-invariant network~(TINet) based on the Strong Baseline to account for the large orientation variation. TINet only requires adding additional consistency losses between the classification and regression outputs of original and transformed images. 

\item We reproduced multiple generic FSOD methods for FSOD on RSIs and created an extensive benchmark for follow-up works on FSOD on RSIs. These reproduced generic methods exhibit strong performance even compared with recent FSOD methods dedicated to RSIs.

\end{itemize}

\section{Related Work}

\subsection{Few-shot Object Detection}
Few-shot Object Detection (FSOD) can be classified into two main approaches: meta-learning-based and transfer-learning-based methods. Meta-learning-based approaches, such as FSRW \cite{kang2019}, aim to extract generalized knowledge across different tasks by learning to learn. These approaches have been extended to a two-stage network, specifically Faster-RCNN, by subsequent works \cite{yan2019, fsdetview, fan2020}, resulting in significant accuracy improvements. On the other hand, transfer-learning-based approaches follow a two-phase strategy. They are initially trained on instances of base categories and then fine-tuned on a limited number of base and novel samples. TFA \cite{wang2020few} improves the fine-tuning process by employing a cosine similarity-based classifier to fine-tune the last layer of Faster-RCNN. FSCE \cite{fsce} addresses misclassification issues by introducing contrastive learning based on TFA. DeFRCN \cite{defrcn} and CFA \cite{cfa} enhance network performance by focusing on loss gradients. In this paper, we focus on the meta-learning-based method. However, we have observed that in the context of RSIs, conventional meta-learning-based approaches fail to achieve comparable performance to transfer-learning-based methods. This discrepancy can be attributed to the fact that meta-learning-based approaches only utilize the C4 layer for RoI pooling, while transfer-learning-based methods employ a Feature Pyramid Network (FPN) to enhance multi-scale feature extraction. To address this gap, we naturally incorporate FPN into the support branch of meta-learning-based methods. Additionally, we introduce depth-wise convolution to emphasize the aggregation between support features and query features. These operations enable us to establish a Strong Baseline that achieves results comparable to transfer-learning-based methods in remote sensing images.

\subsection{FSOD in Remote Sensing Images}
Compared to natural semantic images, RSIs exhibit a greater diversity in the size and orientation of objects. To address these challenges, previous works in the field have introduced more advanced feature extraction modules for adapting FSOD to RSIs \cite{Li2022, Zhao2022, Zhou2022, Xiao2021, zhang2022few}. Additionally, researchers have approached this problem from different perspectives. For instance, Cheng et al. \cite{Cheng2022} proposed a prototype-guided Region Proposal Network (RPN) that incorporates support feature information into candidate box scores, enabling better region proposal generation. Zhang et al. \cite{Zhang2021} employed oriented augmentation of support features to alleviate the diversity in object orientation. In contrast to these existing approaches, our method aims to improve network accuracy without compromising speed by avoiding the introduction of excessive feature extraction modules. Instead, we propose a simple modification to the network architecture by incorporating an FPN and depth-wise convolution. This modification enhances the network's capability to detect objects of diverse scales. Furthermore, to handle the diverse orientation of objects, we propose a transformation-invariant network that encourages the model to be invariant to transformations applied to input images.

\begin{figure}[tp]
	\centering
\resizebox{0.95\linewidth}{!}{\input{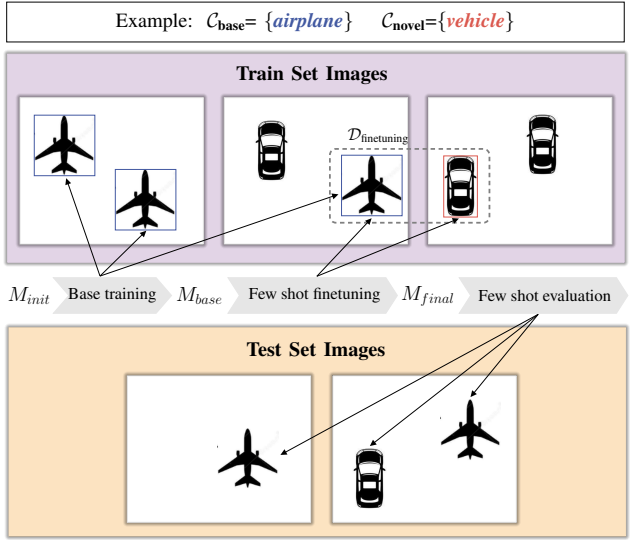}}
	\caption{Example of 2-way 1-shot few-shot object detection. The training processing can be divided into two phases (base training and few-shot finetuning). In base training, the init model $M_{init}$ is trained on base classes (airplane), resulting in $M_{base}$. Then $M_{base}$ can be transferred to $M_{final}$ through fine-tuning on $\mathcal{D}_{\text {finetuning}}$, which includes one sample from each of the novel (vehicle) and base (airplane) classes. In the phase of evaluation, the $M_{final}$ is applied to all the objects in the test set, which includes both novel and base classes. 
	}
	\label{fig:problem_setting}
\end{figure}

\subsection{Transformation Invariant Learning}
Transformation invariant learning has been widely adopted in various domains, including natural images \cite{meanteachers, tip, laine2016temporal}, remote sensing images \cite{feng2022weakly, feng2021saenet, redet, ricnn}, and other scenes \cite{semicurv, xu2020weakly, marcos2017rotation}, which aims to enforce invariance within neural networks. There are two kinds of methods employed to achieve this objective: making the convolutional layer invariant \cite{redet, ricnn}, or enforcing invariance through the loss function \cite{tip, feng2022weakly, semicurv}. In this paper, we focus on the latter approach. This choice is driven by the fact that the former approach necessitates a substantial amount of data to enable the feature learning process to capture invariant features, which can be unrealistic in the few-shot setting. In the area of FSOD, TIP \cite{tip} introduces consistency regularization on predictions from various transformed images. However, its consideration is limited to classification consistency between two augmentations, restricting the use of non-geometric transformation methods (e.g., Gaussian noise and cutout) on input images. For remote sensing object detection tasks, regression consistency enforces the spatial locations to be consistent and should also be considered. Different from the above methods, our method incorporates this idea into FSOD in RSIs and verifies the influence of the different transformations and regularizations on the results. It is obvious that our method is the first attempt to address the obstacle of transformed variations in RSIs under few-shot settings.

\section{Proposed Methods}
\begin{figure*}[tp]
	\centering
\resizebox{0.95\linewidth}{!}{\input{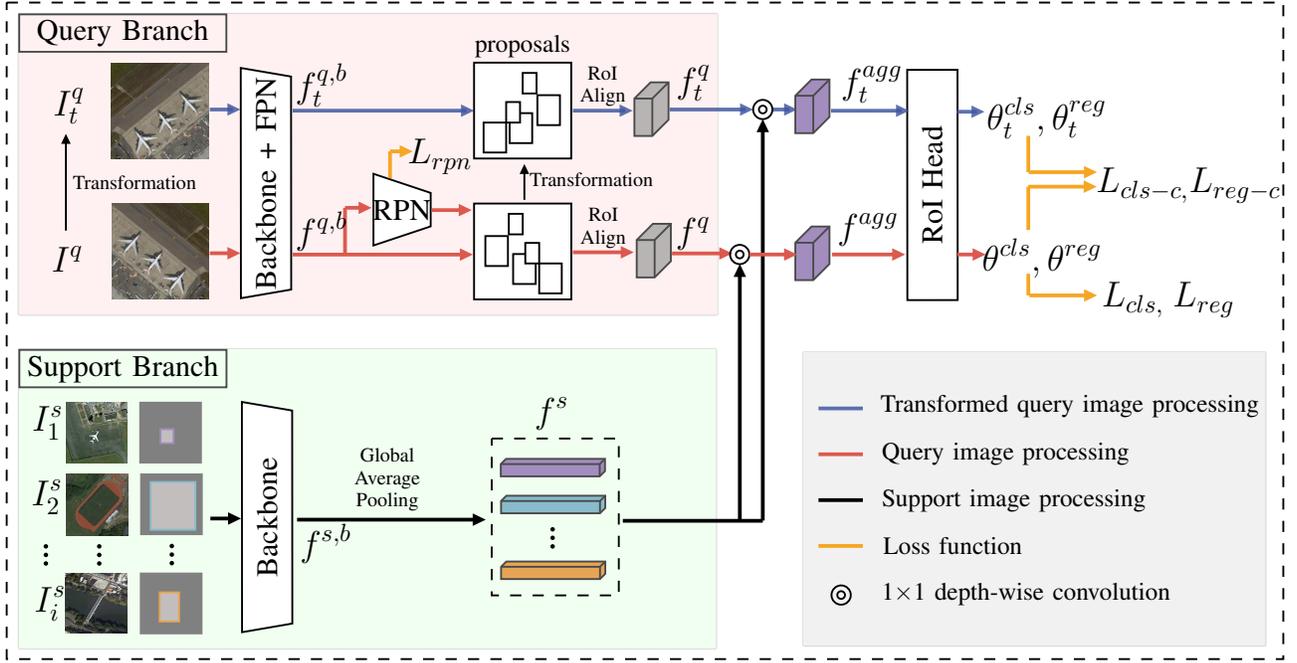}}
	\caption{
	The overall architecture of TINet. The input image $I^q$ and its transformed version $I^q_t$ are fed into a shared backbone network and FPN to obtain query feature map $f^{q,b}$ and $f^{q,b}_t$. After subsequent processing, such as proposal generation by region proposal network (RPN) and the RoI Align, $f^{q,b}$ and $f^{q,b}_t$ are aggregated with the support feature $f^s$, which is generated by the support branch. The aggregated features $f^{agg}_t$ and $f^{agg}_t$ are fed to the RoI Head to obtain regression parameters ($\theta^{reg}$ and $\theta^{reg}_t$) and classification scores ($\theta^{cls}$ and $\theta^{cls}_t$). The supervision loss of the entire network consists of the consistency loss ($L_{cls-c}$, $L_{reg-c}$) and the losses of the Faster-RCNN ($L_{rpn}$,$L_{cls}$ and $L_{reg}$). In the testing phase, the transformed image processing branch is not used. 
	}
	\label{fig:TINet}
\end{figure*}

\subsection{Problem Setting}

As in the previous works \cite{yan2019,kang2019,fsdetview}, we follow the standard problem settings of meta-learning-based FSOD in our paper. Specifically, the required data can be divided into two sets of categories, $\mathcal{C}_{\text {base}}$ and $\mathcal{C}_{\text {novel}}$, where $\mathcal{C}_{\text {base}} \cap \mathcal{C}_{\text {novel}}=\varnothing$. The few-shot object detector aims at detecting objects of $\mathcal{C}_{\text {base}} \cup \mathcal{C}_{\text {novel}}$ by learning from a base dataset $\mathcal{D}_{\text {base}}$ with abundant annotated objects of $\mathcal{C}_{\text {base}}$ and a novel dataset $\mathcal{D}_{\text {novel}}$ with very few annotated objects of $\mathcal{C}_{\text {novel}}$. In the task of $K$-shot object detection, there are exactly $K$ annotated objects for each novel class in $\mathcal{D}_{\text {novel}}$. For the meta-learning approaches, the detector is trained in two phases, i.e., base training and few-shot fine-tuning. In the first phase, the init model $\mathcal{M}_{\text {init}}$ is trained to base model $\mathcal{M}_{\text {base}}$ only using base dataset $\mathcal{D}_{\text {base}}$. An episodic training scheme is applied, where each of the $E$ episodes mimics the $N$-way $K$-shot setting. In each episode $e$, the model is trained on $K$ training examples of $N$ categories on a random subset $\mathcal{D}_{\text {meta}}^{e} \subset \mathcal{D}_{\text {base}},\left|\mathcal{D}_{\text {meta}}^{e}\right|=K \cdot N$. Then, in the few-shot fine-tuning phase, the base model $\mathcal{M}_{\text {base}}$ also adopts the same episodic training scheme as in the first phase, resulting in the final model $\mathcal{M}_{\text {final}}$. Different from the dataset $\mathcal{D}_{\text {base}}$, the fine-tuning dataset $\mathcal{D}_{\text {finetuning}}$ contains $K$ training examples from each of the categories in both the novel and base datasets. Hence, the entire training process can be simply expressed as follows:

\begin{equation}
\mathcal{M}_{\text {init}} \underset{e=1 \ldots E}{\stackrel{\mathcal{D}_{\text {meta }}^{e} \subset \mathcal{D}_{\text {base }}}{\xrightarrow{\hspace*{1.3cm}}}} 
\mathcal{M}_{\text {base}} \stackrel{\mathcal{D}_{\text {finetune }}}{\xrightarrow{\hspace*{1.3cm}}} \mathcal{M}_{\text {final}}.
\end{equation}
In the phase of few-shot evaluation, the final model $\mathcal{M}_{\text {final}}$ is applied to test datasets that contain objects from both the novel and base categories. \figref{fig:problem_setting} provides a visual representation of this process.

\begin{table}[]
\footnotesize
\renewcommand\arraystretch{1.0}
\centering
\caption{Different analysis of Strong Baseline on DIOR test set in split1 (nAP). 20 and 30 means in the setting of 20-shot and 30-shot, respectively. The first row without any strategy is equivalent to Meta-RCNN~\cite{yan2019}. With the addition of different strategies, the results are continuously improved.}
\begin{tabular}{c|c|c|c|c}
\toprule
 \multicolumn{1}{c|}{\begin{tabular}[c]{@{}c@{}}FPN\end{tabular}} & \multicolumn{1}{c|}{\begin{tabular}[c|]{@{}c@{}} Depth-wise\\ convolution\end{tabular}}&\multicolumn{1}{c|}{\begin{tabular}[c|]{@{}c@{}} IoU threshold\\ increasing\end{tabular}} & 20& 30 \\ 
\midrule
 &&&27.9&30.0\\
\CheckmarkBold&&&30.5&34.3\\ 
\CheckmarkBold&\CheckmarkBold&&31.5&36.0   \\ 
\CheckmarkBold&\CheckmarkBold&\CheckmarkBold&\textbf{32.1}&\textbf{36.8}\\ 
\bottomrule
\end{tabular}
\label{tab:strong_baseline_ab}
\end{table}

\subsection{Strong Baseline}
\label{sec:strong Baseline}

The most meta-based few-shot object detection algorithms~\cite{vfa,fsdetview,yan2019} are built upon Faster-RCNN~\cite{ren2015} which uses the backbone's C4 layer features for detection. Because the size variation of objects in natural images is small. So only using the C4 layer of the backbone can play a very good role. However, objects in RSIs often experience a larger variation in scale. Thus, we are motivated to incorporate multi-level feature extraction to account for the scale variation. The most intuitive idea is to add feature pyramid network~(FPN)~\cite{lin2017} to the few-shot detector e.g. Meta-RCNN~\cite{yan2019}, which has a query branch and a support branch. We only add the FPN to the query branch because complex feature fusion problems will be involved when the support branch produces multiple features. As shown in Tab.~\ref{tab:strong_baseline_ab}, when FPN is added to Meta-RCNN, the performance is greatly improved, suggesting the importance of handling large-scale variation. Prior works \cite{vfa,fsdetview,yan2019} resize both the query feature and support feature to a size of 1 $\times$ 1 for element-wise multiplication. In our case, we only process the support feature, resizing it to 1 $\times$ 1 while keeping the query feature unchanged. This allows the RoI head to obtain more information for accurate object identification. Aggregation operation is then performed using depth-wise convolution. Additionally, during the test phase, we increase the IoU threshold in the non-maximum suppression (NMS) step of the region proposal network (RPN) from 0.7 to 0.9. This adjustment helps prevent the removal of bounding boxes due to mistakes, particularly for novel categories. From Tab.~\ref{tab:strong_baseline_ab}, it can be observed that the addition of depth-wise convolution and the increased IoU threshold slightly improve the results.

\subsection{Transformation-Invariant Network}
\label{sec:TINet}
The proposed Strong Baseline effectively tackles the challenge of handling significant scale changes in objects. However, it falls short of addressing the issue of varying object orientations. Although data augmentation can mitigate this problem by introducing orientation transformations to the input image, it fails to address the inconsistency in aggregation between query features and support features. To overcome this limitation, we introduce a transformation-invariant few-shot object detection network (TINet) based on the Strong Baseline. The overall architecture of TINet is illustrated in Fig.~\ref{fig:TINet}. In the subsequent sections, we provide a comprehensive explanation of the query branch, support branch, feature aggregation, loss designs, and testing procedure employed in TINet.

\subsubsection{Query branch}
Given a query image $I^q \in \mathbb{R}^{C \times H \times W}$, we first generate its transformed version $I^q_t \in \mathbb{R}^{C \times H \times W}$. Subsequently, both $I^q$ and $I^q_t$ are passed through a shared backbone network and Feature Pyramid Network (FPN) to produce feature maps $f^{q,b}$ and $f^{q,b}_t$, respectively. To ensure consistency in the generated proposals, only $f^{q,b}$ is utilized as input for the Region Proposal Network (RPN). The same transformation is applied to generate transformed proposals for $I^q_t$. Finally, RoI features $f^{q} \in \mathbb{R}^{C_q \times H_q \times W_q}$ and $f^{q}_t \in \mathbb{R}^{C_q \times H_q \times W_q}$ are extracted using the RoI Align operation, with $H_q$ and $W_q$ set to 7.

\subsubsection{Support branch}
Similar to previous works \cite{yan2019, kang2019}, the support branch takes 4-channel N-way K-shot support images $I_i^s$ as input, consisting of an RGB image and its binary mask derived from the object bounding box. The support feature maps $f^{s} \in \mathbb{R}^{C_s \times 1 \times 1}$ are obtained through backbone feature extraction and global average pooling (GAP) operation.

\subsubsection{Feature aggregation}
For feature aggregation, we employ a 1 $\times$ 1 depth-wise convolution, which is both simple and effective. Specifically, the support feature $f^{s} \in \mathbb{R}^{C_s \times H_s \times W_s}$ serves as a convolution kernel for a 1 $\times$ 1 depth-wise convolution. The resulting aggregated features, denoted as $f^{agg}$ and $f^{agg}_t$, are then fed into the RoI Head to obtain classification scores $\theta^{cls}$, $\theta^{cls}_t$, and regression parameters $\theta^{reg}$, $\theta^{reg}_t$.

\subsubsection{Consistency loss}

By minimizing the consistency loss, the network can perform consistency detection results on both query image processing and its transformation version so that the aggregation feature $f^{agg}$ and $f^{agg}_t$ are forced to the same distribution. The consistency loss comprises two components: the classification consistency loss $L_{cls-c}$ and the regression consistency loss $L_{reg-c}$. Let $\sigma^j$ and $\sigma^j_t$ denote the classification distribution for the $j$-th proposal in $\theta^{cls}$ and $\theta^{cls}_t$, respectively. Different metrics, such as $L_2$ loss, Jensen‒Shannon divergence (JSD), and Kullback‒Leibler divergence (KLD), can be used to measure the distance between these distributions. Through experiments, we find that $L_2$ loss performs the best (as discussed in Section \ref{sec:regularization}). Therefore, the classification consistency loss can be defined as follows:
\begin{equation}
L_{cls-c} = \sum_{j=1}^J\|\sigma^j - \sigma^j_t\|^2
\end{equation}
In contrast to the classification distribution, regression parameters change with image transformation. Let $[\Delta x^j, \Delta y^j, \Delta w^j, \Delta h^j]$ represent the regression results for the $j$-th proposal in $\theta^{reg}$, representing the offset of the center point and the scale coefficients for width and height. Similarly, $[\Delta x^j_t, \Delta y^j_t, \Delta w^j_t, \Delta h^j_t]$ represents the $j$-th parameters in $\theta^{reg}_t$. In this paper, we focus on the effect of three flipping transformations: horizontal flipping, vertical flipping, and diagonal flipping. Since the flipping transformation causes $\Delta x^j_t$ or $\Delta y^j_t$ to move in the opposite direction, a negative operation is applied to correct it. For example, in diagonal flipping, $\Delta x^j$ and $\Delta y^j$ correspond to $- \Delta x^j_t$ and $- \Delta y^j_t$. The regression consistency loss with $L_2$ regularization can be defined as:
\begin{equation}
  \begin{aligned}
  L_{reg-c} = & \sum_{j=1}^J\left[\left\|\Delta x^j \!- \!( \!- \!\Delta x^j_t)\right\|^2 \!+ \!\left\|\Delta y^j \!- \!( \!- \!\Delta y^j_t)\right\|^2\right. \\
   &\left. \!+ \!\left\|\Delta w^j \!- \!\Delta w^j_t\right\|^2 \!+ \!\left\|\Delta h^j \!- \!\Delta h^j_t\right\|^2\right]
  \end{aligned}
\end{equation}
For the other two flipping transformations, $\Delta x^j$ and $\Delta y^j$ should correspond to $- \Delta x^j_t$ and $\Delta y^j_t$ for horizontal flipping and $\Delta x^j_t$ and $- \Delta y^j_t$ for vertical flipping. The complete procedure to compute the consistency loss is presented in Algorithm \ref{alg:loss_alg}. Both the base training phase and the few-shot fine-tuning phase utilize the consistency loss.

\begin{algorithm}

\caption{Consistency loss computation}
\label{alg:loss_alg}
  \begin{algorithmic}[1]
    \renewcommand{\algorithmicrequire}{\textbf{Input:}}
    \renewcommand{\algorithmicensure}{\textbf{Output:}}
    \REQUIRE Query image $I^q$ and its transformed version $I^q_t$, support images $I^s_1,...,I^s_i$.
    \ENSURE Consistency loss $L_{reg-c}$, $L_{cls-c}$. \\

\textbf{Query Branch:} 
    \STATE \quad Feed $I^q$ and $I^q_t$ to the shared backbone and FPN to \\
    \quad obtain output features $f^{q, b}$ and $f_t^{q, b}$;
    \STATE \quad Generate proposals using $f^{q, b}$ by RPN and obtain the \\
    \quad transformed proposals;
    \STATE \quad Extract query box feature $f^q_t$ and $f^q$ by RoI Align.\\
    \textbf{Support Branch:}
    \STATE \quad Feed $I^s_1,...,I^s_i$ to the shared backbone to obtain output\\
    \quad feature $f_1^{s,b},...,f_i^{s,b}$;
    \STATE \quad Generate support features $f^s$ by GAP;
    \STATE \quad Random sample a support feature $f^s_r$ from $f^s$.\\
    \STATE $f^{agg}_t \leftarrow f^{q}_t \otimes f^s_r$;
    \quad $f^{agg} \leftarrow f^{q} \otimes f^s_r$;
    \STATE Feed $f^{agg}_t$ and $f^{agg}_t$ to RoI head to obtain $\theta^{cls}$ and $\theta^{reg}$, $\theta^{cls}_t$ and $\theta^{reg}_t$;
    \STATE Compute $L_{cls-c}$ according to Eq. (2)
    and $L_{reg-c}$ according to Eq. (3).
  \end{algorithmic}

\end{algorithm}

\subsubsection{Total loss}
We optimize the total loss $L_T$ by combining the standard supervised loss terms with the consistency losses, as follows:
\begin{equation}
L_T = L_{rpn} + L_{cls} + L_{reg} + \lambda_{1} L_{cls-c} + \lambda_{2} L_{reg-c}
\end{equation}
Here, $L_{rpn}$, $L_{cls}$, and $L_{reg}$ represent the losses (cross-entropy loss and L1 loss) for the Region Proposal Network (RPN), the classification loss and regression loss, respectively. The terms $\lambda_1$ and $\lambda_2$ control the strength of the transformation consistency. As object detection consistency is not stable in the early training stage, we choose relatively smaller weights: $\lambda_1 = 0.05$ and $\lambda_2 = 0.02$.

\subsubsection{Testing procedure}
During training, the support feature is randomly selected for each iteration. However, during testing, all support features must be involved in the process. The testing phase is outlined in Algorithm \ref{alg:test_alg}. Importantly, the transformed images are not included in the testing phase, ensuring that there is no impact on the overall inference time. The impact of different components in the consistency loss on both training and testing time is discussed in Section \ref{sec:training and testing time}.

\begin{algorithm}[!htb]

  \caption{The overall test procedure (one image)}
     \label{alg:test_alg}
  \begin{algorithmic}[1]
    \renewcommand{\algorithmicrequire}{\textbf{Input:}}
    \renewcommand{\algorithmicensure}{\textbf{Output:}}
    \REQUIRE Query image $I^q$ and support images $I^s_1,...,I^s_i$.
    \ENSURE Classification score $\theta^{cls}$, regression parameters $\theta^{reg}$. \\ 

\textbf{Query Branch:}
    \STATE \quad Feed $I^q$ to the shared backbone and FPN to obtain \\ 
    \quad output feature $f^{q, b}$;
    
    \STATE \quad Generate proposals using $f^{q, b}$ by RPN;
     
    \STATE \quad Extract query box feature $f^q$ by RoI Align.\\

\textbf{Support Branch:}
    \STATE \quad Feed $I^s_1,...,I^s_i$ to shared backbone to obtain output \\
    \quad feature $f_1^{s,b},...,f_i^{s,b}$;
    \STATE \quad Generate support features $f^s$ by GAP.
    \FOR{$i \leftarrow 1$ to $N$}    
    \STATE $f^{agg}_i \leftarrow f^{q} \otimes f^s_i$;
    \STATE Feed $f^{agg}_i$ to RoI head to obtain $\theta^{cls}_i$ and $\theta^{reg}_i$.
    \ENDFOR

  \end{algorithmic}

\end{algorithm}



\begin{table*}[!htb]
    \renewcommand\arraystretch{1.0}
    \caption{Different novel/base classes split settings for our experiments.}
    \centering
    \footnotesize
    \begin{tabular}{p{3.0cm}<{\centering}|p{9.0cm}<{\centering}|p{2cm}<{\centering}}
        \toprule
        Dataset(split)&Novel classes&Base classes \\
        \midrule
DIOR(split1)   & airplane, baseball field, train station, tennis court, windmill        & rest         \\ \midrule
DIOR(split2)   & airplane, airport, expressway toll station, harbor, ground track field & rest         \\ \midrule
HRRSD          & airplane, baseball diamond, ground track field, storage tank           & rest         \\ \midrule
NWPU VHR-10.v2 & airplane, baseball diamond, tennis court                                   & rest         \\ 

        \bottomrule
    \end{tabular}
    \label{tab:split}
\end{table*}

\begin{table*}

    \renewcommand\arraystretch{1.0}
	\caption{  
    Comparing results of different few-shot object detection methods on the DIOR test set in \textbf{split1}(nAP/bAP/mAP). Colored results represent the \textcolor{red}{best} and \textcolor{blue}{second-best}. $^\ast$ indicates results reported in \cite{Li2022}.
     }
	\footnotesize
	\centering
	\begin{tabular}{c|c|c|ccc|ccc|ccc|ccc} 

		\toprule
		Method&Backbone&Combination&\multicolumn{3}{c|}{5-shot}&\multicolumn{3}{c|}{10-shot}&\multicolumn{3}{c|}{20-shot}&\multicolumn{3}{c}{30-shot}\\
		\midrule

        & & & nAP & bAP & mAP & nAP & bAP & mAP &
        nAP & bAP & mAP & nAP & bAP & mAP  
        \\
        FRCN-ft \cite{ren2015} &ResNet-50&FPN&15.9&33.7&29.3&20.4&43.4&37.7 & 24.8& 49.4&43.2&26.5&50.7&44.7\\
        FsDetView \cite{fsdetview}&ResNet-50&C4&17.0&37.8&32.6&21.9&39.7&35.3 & 24.9& 41.8&37.6&27.6&46.7&41.9\\
        TFA \cite{wang2020few}&ResNet-50&FPN&21.9&56.1&47.6&24.1&\textcolor{red}{58.0}&49.5&32.9&56.9&50.9&33.4&58.9&52.5\\
        FSCE \cite{fsce}&ResNet-50&FPN&22.8&\textcolor{blue}{56.9}&48.4&30.3&\textcolor{blue}{57.6}&50.8&33.7&\textcolor{blue}{60.2}&53.5&37.4&60.7&54.9\\
        Meta-RCNN \cite{yan2019}&ResNet-50&C4&20.7&47.1&40.5&24.7&46.7&41.3&27.9& 48.1&43.0&30.0&49.3&44.5\\
        RepMet$^\ast$ \cite{repmet}&InceptionV3&-&8.0&-&-&14.0&-&-&16.0& -&-&-&-&-\\
        FSRW$^\ast$ \cite{kang2019}&Darknet-19&-&22.0&-&-&28.0&-&-&34.0& -&-&-&-&-\\
        FSODM$^\ast$ \cite{Li2022}&Darknet-53&-&25.0&-&-&32.0&-&-&36.0& -&-&-&-&-\\
        Zhang et al. \cite{zhang2022few} &ResNet-101&FPN&\textcolor{red}{34.0}&-&-&\textcolor{blue}{37.0}&-&-&\textcolor{blue}{42.0}& -&-&-&-&-\\ 
        \midrule
        Strong Baseline~(Ours)&ResNet-50&FPN&22.0&48.0&41.5&26.9&52.1&45.8&32.1& 55.5&49.6&36.8&55.4&50.7\\

        TINet~(Ours)&ResNet-50&FPN&\textcolor{blue}{29.5}&56.2&\textcolor{blue}{49.5}&35.2&56.8&\textcolor{blue}{51.4}&41.6& 59.8&\textcolor{blue}{55.3}&\textcolor{blue}{42.8}&\textcolor{blue}{62.6}&\textcolor{blue}{57.7}\\
        TINet~(Ours)&ResNet-101&FPN&28.6&\textcolor{red}{57.8}&\textcolor{red}{50.5}&\textcolor{red}{38.4}&57.4&\textcolor{red}{52.7}&\textcolor{red}{43.2}&\textcolor{red}{ 62.1}&\textcolor{red}{57.4}&\textcolor{red}{44.6}&\textcolor{red}{63.6}&\textcolor{red}{58.9}\\

	\bottomrule
		  
	\end{tabular}
    \label{tab:dior_split1_exp}
\end{table*}

\begin{table}

    \renewcommand\arraystretch{1.0}
	\caption{Comparing results of different FSOD methods on the DIOR test set in \textbf{split2} (nAP). Colored results represent the \textcolor{red}{best} and \textcolor{blue}{second-best}. $^\ast$ indicates results reported in \cite{Cheng2022},
     }
    \footnotesize
	\centering
        \setlength{\tabcolsep}{1.8mm}{
	\begin{tabular}{c|c|c|c|c|c} 
		\toprule
		Method&Backbone&5&10&20&30\\
		\midrule

        FRCN-ft \cite{ren2015} &ResNet-50&14.9&17.6&22.8&23.5\\
        FsDetView \cite{fsdetview}&ResNet-50&14.2&16.2&19.1&21.9\\
        TFA \cite{wang2020few}&ResNet-50&18.0&20.9&23.0&26.4\\
        FSCE \cite{fsce} &ResNet-50&19.9&22.7&26.9&30.6\\
        Meta-RCNN \cite{yan2019}&ResNet-50&14.1&17.6&21.0&21.2\\
        RepMet$^\ast$ \cite{repmet}  &InceptionV3&5.6&5.9&6.8&6.5\\
        FSRW$^\ast$ \cite{kang2019}  &DarkNet-19&7.0&9.0&14.1&14.4\\
        P-CNN$^\ast$ \cite{Cheng2022} &ResNet-101&14.9&18.9&22.8&25.7\\
        Zhang et al. \cite{zhang2022few}&ResNet-101&15.5&19.7&23.8&29.6\\
        G-FSDet \cite{gfsod}&ResNet-101&15.8&20.7&22.7&-\\
        
        \midrule
        Strong Baseline&ResNet-50&20.1&23.3&26.5&28.1\\
        TINet&ResNet-50&\textcolor{blue}{21.7}&\textcolor{blue}{24.1}&\textcolor{blue}{28.0}&\textcolor{blue}{31.9}\\
        TINet&ResNet-101&\textcolor{red}{22.8}&\textcolor{red}{25.1}&\textcolor{red}{29.4}&\textcolor{red}{33.2}\\
		\bottomrule
	\end{tabular}}
    \label{tab:dior_split2_exp}
\end{table}

\begin{table*}

    \renewcommand\arraystretch{1.0}
	\caption{Comparison result of different few-shot object detection methods on HRRSD test set (nAP/bAP/mAP). Colored results represent the \textcolor{red}{best} and \textcolor{blue}{second-best}.}
	\footnotesize
	\centering
	\begin{tabular}{c|c|c|ccc|ccc|ccc|ccc} 
		\toprule
		Method&Backbone&Combination&\multicolumn{3}{c|}{5-shot}&\multicolumn{3}{c|}{10-shot}&\multicolumn{3}{c|}{20-shot}&\multicolumn{3}{c}{30-shot}\\
		\midrule
        & & & nAP & bAP & mAP & nAP & bAP & mAP &
        nAP & bAP & mAP & nAP & bAP & mAP  
        \\
        FRCN-ft~\cite{ren2015} &ResNet-50&FPN&26.9&\textcolor{blue}{79.4}&63.3&38.1&\textcolor{red}{80.8}&67.7&44.0&\textcolor{red}{82.1}&70.4&46.2&\textcolor{red}{82.9}&71.6\\
        FsDetView~\cite{fsdetview}&ResNet-50&C4&35.6&62.2&54.0&42.0&67.8&59.8&48.1&69.8&63.1&52.8&70.0&64.7\\
        TFA~\cite{wang2020few}&ResNet-50&FPN&36.0&75.3&63.2&45.1&79.3&68.8&51.4&80.7&71.7&53.0&\textcolor{blue}{81.0}&72.4\\
        FSCE~\cite{fsce} &ResNet-50&FPN&\textcolor{blue}{37.5}&75.7&\textcolor{blue}{64.0}&\textcolor{blue}{46.3}&79.8&\textcolor{blue}{69.5}&\textcolor{blue}{54.5}&\textcolor{blue}{80.9}&\textcolor{blue}{72.8}&\textcolor{blue}{61.9}&80.6&\textcolor{blue}{74.9}\\
        Meta-RCNN~\cite{yan2019}&ResNet-50&C4&30.5&71.1&58.6&41.1&73.2&63.3&47.7&73.7&65.7&51.5&75.5&68.1\\
       \midrule
        Strong Baseline&ResNet-50&FPN&32.3&71.2&59.2&43.7&75.4&65.6&53.3&75.1&68.4&61.4&77.2&72.3\\
        TINet&ResNet-50&FPN&\textcolor{red}{38.3}&\textcolor{red}{81.8}&\textcolor{red}{68.4}&\textcolor{red}{47.3}&\textcolor{blue}{80.2}&\textcolor{red}{70.4}&\textcolor{red}{58.9}&80.5&\textcolor{red}{73.9}&\textcolor{red}{64.3}&80.8&\textcolor{red}{75.3}\\
        \bottomrule
	\end{tabular}
    \label{tab:hrrsd_exp}
\end{table*}

\begin{table}
    \renewcommand\arraystretch{1.0}
	\caption{Comparison result of different FSOD methods on NWPU VHR-10.v2 test set (nAP). Colored results
    represent the \textcolor{red}{best} and \textcolor{blue}{second-best}.}
    \footnotesize
	\centering
        \setlength{\tabcolsep}{1.8mm}{
	\begin{tabular}{c|c|c|c|c|c} 
		\toprule
		Method&Backbone&2&3&5&10\\
		\midrule

        FRCN-ft \cite{ren2015} &ResNet-50&35.1&44.8&48.9&57.3\\
        FsDetView \cite{fsdetview}&ResNet-50&40.8&52.2&58.6&65.2\\
        TFA \cite{wang2020few}&ResNet-50&42.8&50.7&53.1&60.5\\
        FSCE \cite{fsce} &ResNet-50&\textcolor{blue}{53.4}&\textcolor{red}{56.4}&\textcolor{blue}{60.6}&\textcolor{blue}{68.7}\\
        Meta-RCNN \cite{yan2019}&ResNet-50&43.1&50.6&55.1&62.6\\
        OFA  \cite{Zhang2021} &ResNet-101&34.0&43.2&60.4&66.7\\
        G-FSDet  \cite{gfsod} &ResNet-101&-&49.1&56.1&\textcolor{red}{71.8}\\
        \hline
        Strong Baseline&ResNet-50&45.8&55.1&59.8&64.0\\
        TINet&ResNet-50&\textcolor{red}{53.7}&\textcolor{blue}{55.8}&\textcolor{red}{63.5}&\textcolor{red}{71.8}\\

		\bottomrule
	\end{tabular}}
    \label{tab:nwpu_exp}
\end{table}

\section{Experiments}
\label{sec:experiments}
\subsection{Dataset and Experimental Setting}
\label{sec:setting}
We conduct experiments on three extensively used remote sensing datasets, i.e., NWPU VHR-10.v2 \cite{li2017}, DIOR \cite{li2020} and HRRSD \cite{hrrsd}. NWPU VHR-10.v2 contains 1172 annotated images distributed into ten categories, which are divided into 75\% for training and 25\% for testing. For the DIOR dataset, 11,725 images are used as the training set, and the remaining 11,738 images are employed as the test set. Likewise, the HRRSD data are divided into three parts (the training, validation, and test sets), with 5,401, 5,417, and 10,913 images, respectively.

To establish the few-shot learning setup, we further divide each dataset into two parts, the novel class and the base class following the practice adopted in~\cite{Li2022}\cite{Cheng2022}. A detailed split setting is presented in Tab.~\ref{tab:split}. It should be noted that the number of shots denotes the number of instances that are not images because one image contains several instances. We evaluate the testing images, which contain both base and novel classes.

For all experiments conducted with our proposed detector, we utilized a ResNet \cite{resnet} backbone network pre-trained on ImageNet. During the training process, we employed an SGD optimizer with a momentum coefficient of 0.9 and a weight decay of 0.0001. The batch size was set to 4 for all datasets. In the base training stage, the initial learning rate was set to 0.01, with a 0.1 decrease at 80\% of the total iterations. In the few-shot fine-tuning stage, the initial learning rate was set to 0.001, with a 0.1 decrease at 80\% of the total iterations. For NWPU VHR-10.v2, we trained for 9,000 iterations in the base training stage and 3,000 iterations in the few-shot fine-tuning stage. For HRRSD and DIOR, we trained for 36,000 iterations in the base training stage and 6,000 iterations in the few-shot fine-tuning stage. Additionally, we employed multiscale training and random flipping to enhance the detection performance. The scale range of the input images varies (440, 472, 504, 536, 568, 600). We perform the experiments under the PyTorch framework on a PC with an Intel single-core i7 CPU and a GeForce RTX 3090 GPU.

In the subsequent experimental results, we adopt the evaluation protocol of the PASCAL visual object classes (VOC) \cite{pascal}. The mean average precision (mAP) represents the average precision across all object categories, including both base and novel categories. The novel class average precision (nAP) indicates the average precision for the novel categories, while the base class average precision (bAP) indicates the average precision for the base categories.

\subsection{Reproducing Generic FSOD Methods}
\label{sec:comparison}
To make a fair comparison, we first reproduce several state-of-the-art generic few-shot object detection methods based on the open-source framework MMFewShot \cite{mmfewshot} which is tailored for few-shot learning. The reproduced methods include FRCN-ft \cite{ren2015}, TFA \cite{wang2020few}, FSCE \cite{fsce}, Meta-RCNN \cite{yan2019}, FsDetView \cite{fsdetview}, as well as our proposed Strong Baseline and TINet. Specifically, the Strong Baseline is referred to in \sectionref{sec:strong Baseline}. FRCN-ft only uses base class objects to train the Faster-RCNN with FPN in the first phase and then uses combinations of the base class and novel class objects to fine-tune in the second phase. For TFA, we only freeze the backbone in the fine-tuning phase because we can obtain better results in this way, which is slightly different from the original paper. For FSCE, Meta-RCNN, and FsDetView, we keep the same setting as the original paper.

\subsection{Few-Shot Object Detection Results}
\label{sec:comparison}
\subsubsection{DIOR}
We present the results of different methods for split1 and split2 in Tab.~\ref{tab:dior_split1_exp} and Tab.~\ref{tab:dior_split2_exp} respectively. 
In split1, as shown in Tab.~\ref{tab:dior_split1_exp}, in addition to the reproduced generic few-shot object detection methods, we further incorporate comparisons with RepMet~\cite{repmet} with InceptionV3 \cite{inception} as the backbone and FSRW with DarkNet-19 \cite{yolo9000} as backbone according to \cite{Li2022}. We make the following observations from the results. First, the TINet outperforms all competing methods except for nAP@5shot. The gap is particularly significant compared with generic FSOD methods where more than $10\%$ improvements in nAP/bAP/mAP are observed throughout 5 to 30~shots. This suggests the strong few-shot learning capability of TINet. Second, all the transfer-learning approaches and meta-learning approaches outperform the original fine-tuned Faster-RCNN (FRCN-ft). For the meta-learning approaches, 
we observe that Meta-RCNN and FsDetView only use the C4 layer for subsequent processing, thus they perform relatively worse compared with the Strong Baseline on the DIOR dataset which features large diversity in object scales. For the transfer-learning-based approaches, both FSCE and TFA are way behind TINet despite all are using FPN, probably due to the large intra-class variation in the DIOR dataset. 
Finally, compared with the results reported by the state-of-the-art methods in RSIs \cite{Li2022}\cite{zhang2022few}, we also demonstrate a competitive result, except for the nAP@5 shots. 
We further present the comparisons for split2 in Tab.~\ref{tab:dior_split2_exp}, which is generally considered to be more challenging than split1. 
Under split2, we observe more significant improvements in TINet from the best-performing methods. These results again validate the effectiveness of the proposed method.

\subsubsection{HRRSD}
The quantitative results obtained by applying different methods to the HRRSD dataset are presented in Table \ref{tab:hrrsd_exp}. Since no previous algorithms have been tested on the HRRSD dataset, we can only compare the results with our implemented algorithm. From the results, we can observe that most methods achieve good performance on this dataset, which is less complex compared to the DIOR dataset and has fewer object categories. Transfer learning approaches, especially FSCE, demonstrate strong performance in certain aspects. However, our method still outperforms FSCE in terms of nAP at 5 shots, 10 shots, 20 shots, and 30 shots, with improvements of 0.8\%, 1.0\%, 4.4\%, and 2.4\%, respectively.

\subsubsection{NWPU VHR-10.v2}
As shown in \tableref{tab:nwpu_exp}, since the NWPU VHR-10.v2 dataset is relatively simple, all the methods achieve relatively good results. It can be observed that our Strong Baseline has higher results than the general meta-learning methods but lower results than the transfer learning approach, FSCE. This is because this dataset is very small and the intra-class similarity is not large. TINet still outperforms FSCE on most metrics. Although OFA
\cite{Zhang2021} improves object recognition in novel categories by rotating the support samples, increases the inference time, and does not use the oriented feature augmentation method in the base training phase, which may reduce the generalization performance.

\subsection{Ablation Study}
We conduct ablation experiments on the DIOR dataset (split1) to reveal the effectiveness of each individual component. Unless otherwise specified, the backbone network chosen is ResNet-50.

\subsubsection{Comparison with data augmentation}
There are two consistency losses ($L_{cls-c}$ and $L_{reg-c}$) in the TINet. As shown in Tab.~\ref{tab:data_augmentation_ab}, we examine the influence of $L_{cls-c}$ and $L_{reg-c}$) and make a comparison with data augmentation, which augments the image before feeding it into the query branch. The experimental results in the first row are obtained without any strategy. It should be noted that for data augmentation, we choose a combination of horizontal, vertical, and diagonal flipping. The consistency loss $L_{cls-c}$ and $L_{reg-c}$ here are both the $L_2$ loss, and the corresponding flipping method is diagonal flipping. It can be observed that data augmentation improves the performance of the network. However, as the number of shots increases, the impact of data augmentation diminishes significantly. On the other hand, adding only the consistency losses $L_{cls-c}$ and $L_{reg-c}$ outperforms the use of data augmentation alone. When either $L_{cls-c}$ or $L_{reg-c}$ is added, the network achieves stable improvements in performance, except in the 5-shot scenario. The best results are obtained when both the consistency losses and data augmentation are used together, indicating that these two techniques are complementary to each other.
\begin{table}[!htb]
\footnotesize
\renewcommand\arraystretch{1.0}

\centering
\caption{Comparison experiments with data augmentation on DIOR test set in split1 (nAP). 5, 10, 20, and 30 mean in the setting of 5-shot, 10-shot, 20-shot, and 30-shot, respectively. Data Aug represents data augmentation. The experimental results in the first row are obtained without any strategy.}
\begin{tabular}{c|c|c|c|c|c|c}
\toprule
Data Aug& $L_{cls-c}$ & $L_{reg-c}$ & 5& 10 & 20 & 30 \\ 
\midrule
 &&&20.8&23.8&30.1&36.3\\
\CheckmarkBold&&&22.0&26.9&32.1&36.8\\ 
&\CheckmarkBold&&29.1&33.9&40.5&40.9   \\ 
&&\CheckmarkBold&27.8&31.8&39.3&41.6   \\ 
&\CheckmarkBold&\CheckmarkBold&28.8&34.3&41.1&42.2  \\ 
\CheckmarkBold&\CheckmarkBold&\CheckmarkBold&\textbf{29.5}&\textbf{35.2}&\textbf{41.6}&\textbf{42.8}\\ 
\bottomrule
\end{tabular}
\label{tab:data_augmentation_ab}
\end{table}

\subsubsection{Alternative transformations}
\label{sec:flipping transformation}
We verify the effect of the different flipping transformations on the experimental results. As shown in Tab.~\ref{tab:flipping_ab}, we observe similar results for both the horizontal and vertical flips. The result of the diagonal flip is slightly better than the previous two. This is because the diagonal flip introduces fewer changes to the object's appearance so that the training is more stable compared to the horizontal and vertical flips.

\begin{table}[!htb]
\footnotesize
\renewcommand\arraystretch{1.0}
\centering
\caption{Detection results of alternative transformations on DIOR test set in split1 (nAP). 5, 10, 20, and 30 means in the setting of 5-shot, 10-shot, 20-shot and 30-shot, respectively.}
\begin{tabular}{c|c|c|c|c}
\toprule
Flipping method& 5 & 10 & 20 & 30 \\ 
\midrule
None&22.0&26.9&32.1&36.8\\
Vertical&28.1&34.1&39.1    &41.3    \\ 
Horizontal&28.5   & 34.5   & 39.4   & 41.2   \\ 
Diagonal&\textbf{29.5}&\textbf{35.2}&\textbf{41.6}&\textbf{42.8}\\  
\bottomrule
\end{tabular}
\label{tab:flipping_ab}
\end{table}

\subsubsection{Alternative consistency regularizations.}
\label{sec:regularization}
We verify the effect of different regularizations in the classification consistency loss $L_{con-c}$ on the results. As shown in Tab.~\ref{tab:regularization_ab}, the JSD and KLD represents
Jensen–Shannon divergence and Kullback–Leibler divergence, respectively. The weight of JSD and KLD here we chose are 0.05 and 0.1. It can be observed that simply using the $L_2$ loss yields the best results in most metrics except in 5-shot. Because the $L_2$ loss is more sensitive to outliers, it can play a more restrictive role.

\begin{table}[!htb]
\footnotesize
\centering
\caption{Detection results with alternative consistency regularizations on DIOR test set in split1 (nAP). 5, 10, 20, and 30 mean in the setting of 5-shot, 10-shot, 20-shot, and 30-shot, respectively. }
\begin{tabular}{c|c|c|c|c}

\toprule
Regularization method & 5 & 10 & 20 & 30 \\ 
\midrule

None&22.0&26.9&32.1&36.8\\
JSD & 27.8  & 34.8   & 38.1   &41.4    \\ 
KLD & \textbf{30.3} &34.7  &  37.9  &40.2    \\ 
$L_2$ &29.5&\textbf{35.2}&\textbf{41.6}&\textbf{42.8}   \\ 
\bottomrule
\end{tabular}
\label{tab:regularization_ab}
\end{table}

\subsubsection{Alternative hyper-parameters of loss $L_T$}
General object detectors always focus on two main sub-tasks (regression and classification) so that the weight of auxiliary losses should be relatively smaller. 
We carried out evaluations on the robustness of the choice of $\lambda$s in Tab.~\ref{tab:lambda_exp}. In general, stable performance is observed around the hyper-parameters we chose.

\begin{table}[!htb]
\footnotesize
	\caption{Comparing different $\lambda$ on DIOR test set in split1 (nAP) under 5, 10, 20, and 30 shots.}

	\centering
        {
	\begin{tabular}{cc|cccc} 
		\toprule
$\lambda_1$&$\lambda_2$&5&10&20&30\\
		\midrule

		1&1&8.6&12.2&16.7&24.8\\
  		0.5&0.5&25.4&31.6&37.7&38.1\\
  		0.05&0.05&29.0&34.7&\textbf{42.2}&42.1\\  
  		0.05&0.02&\textbf{29.5}&\textbf{35.2}&41.6&\textbf{42.8}\\  
  		0.02&0.05&29.2&35.0&41.3&42.4\\  
  		0.02&0.02&29.1&34.7&40.9&41.8\\  
		\bottomrule
	\end{tabular}}
    \label{tab:lambda_exp}
\end{table}

\subsubsection{Pearson correlation coefficient}
\begin{figure}
	
	\centering

	\includegraphics[width=0.9\linewidth]{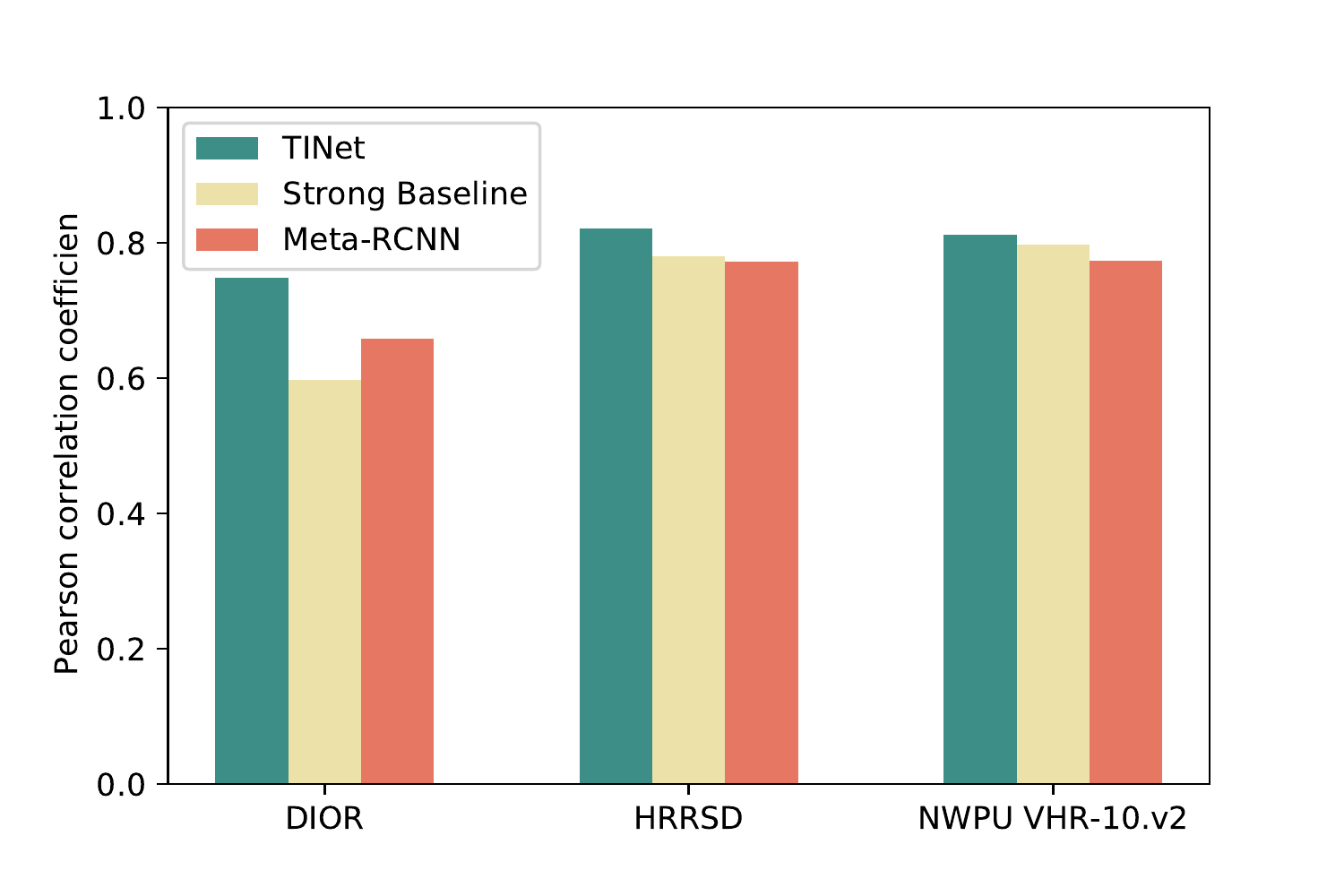}
	
	\caption{Pearson correlation coefficient (PCC) between classification scores and IoUs.}
	\label{fig:pcc_comparison}
\end{figure}

We further measure the calibration of detection models by the Pearson Correlation Coefficient (PCC), defined as follows:

\begin{equation}
   r=\frac{\sum\left(x-m_x\right)\left(y-m_y\right)}{\sqrt{\sum\left(x-m_x\right)^2 \sum\left(y-m_y\right)^2}} 
\end{equation}
$x$ and $y$ are respectively the IoU between ground-truth and predicted bounding boxes and the confidence score (the highest posterior). A high correlation indicates the confidence is well calibrated.
$m_x$ and $m_y$ are the mean of the $x$ and $y$ respectively.
The results (shown in \figref{fig:pcc_comparison}) demonstrate that the TINet obtained a higher value of PCC than the Strong Baseline and Meta-RCNN in all the datasets, suggesting TINet is better calibrated than others.

\subsubsection{Training and testing time}
\label{sec:training and testing time}
\renewcommand\arraystretch{1.0}
The results are shown in Tab.~\ref{tab:time_ab}. Here one iteration includes four multi-scale images, and in the testing phase, the images are resized to 600 $\times$ 600. It can be observed that the training time of TINet increases slightly compared with Strong Baseline because TINet has to process two images simultaneously. However, the inference time of all the comparison methods remains the same. In addition, the loss computation has little effect on the time of network training.
\begin{table}[]
\footnotesize
\centering
\caption{The comparison of training and testing time between the Strong Baseline and TINet.}

\begin{tabular}{c|c|c}
\toprule
\multicolumn{1}{c|}{\begin{tabular}[c]{@{}c@{}} Method\end{tabular}} & \multicolumn{1}{|c|}{\begin{tabular}[c]{@{}c@{}}Training \\(iteration/s)\end{tabular}}& \multicolumn{1}{|c}{\begin{tabular}[c]{@{}c@{}}Testing \\(frame/s)\end{tabular}}\\ 
\midrule

Strong Baseline & 4.92 & 15.5       \\ 
TINet(w/o $L_{cls-c}$) & 4.29 & 15.5     \\ 
TINet(w/o $L_{reg-c}$) & 4.35 & 15.5     \\ 
TINet & 4.27 & 15.5     \\ 
\bottomrule
\end{tabular}
\label{tab:time_ab}
\end{table}

\subsection{Additional analysis}
In this section, we discuss some common issues and the feasibility of alternative methods.

\subsubsection{Why not apply the same transformation to the support branch?}

We considered this approach, but experimental results showed no significant difference in the results compared to the current method (shown in Fig. \ref{tab:support}). Therefore, to ensure training efficiency, we only apply transformations in the query branch. The possible reason for this is that by transforming instances in the query branch while keeping the support instances fixed, the model can learn a sufficient variety of matching methods. In this scenario, adding these transformations in the support branch became unnecessary.

\begin{table}[!htb]

\footnotesize
\centering
\caption{Detection results with applying the transformation to support branch or not on DIOR test set in split1 (nAP). 5, 10, 20, and 30 mean in the setting of 5-shot, 10-shot, 20-shot, and 30-shot, respectively. }
\begin{tabular}{c|c|c|c|c|c}

\toprule
    \multirow{2}*{\makecell[c]{Apply transformation \\to support branch}}&    \multirow{2}*{Backbone}&\multirow{2}*{5}&\multirow{2}*{10}&\multirow{2}*{20}&\multirow{2}*{30}\\ 
    ~&~&~&~&~&~\\
\midrule

\CheckmarkBold&ResNet-101&29.7&35.3&41.8&42.5\\
\XSolidBrush&ResNet-101&29.5&35.2&41.6&42.8\\
\bottomrule
\end{tabular}
\label{tab:support}
\end{table}

\subsubsection{Why remove the Meta-loss?}
Initially, we utilized the meta-loss but later discovered that the results were not better than without using the meta-loss (shown in Tab. \ref{tab:meta_loss}). We hypothesize that the query feature contains both regression and classification information, while the support feature only contains classification information. In our case, although the meta-loss can improve the classification performance of the support feature, it may not necessarily be beneficial for the regression task.

\begin{table}[!htb]
\footnotesize

\centering
\caption{Detection results with meta loss or without meta loss on DIOR test set in split1 (nAP). 5, 10, 20, and 30 mean in the setting of 5-shot, 10-shot, 20-shot, and 30-shot, respectively. }
\begin{tabular}{c|c|c|c|c|c}

\toprule
Method&Backbone& 5 & 10 & 20 & 30 \\ 
\midrule

TINet (w/ Meta-loss)&ResNet-101&27.9&34.1&41.0&41.3\\
TINet (w/o Meta-loss)&ResNet-101&29.5&35.2&41.6&42.8\\
\bottomrule
\end{tabular}
\label{tab:meta_loss}

\end{table}

\subsubsection{Why not apply arbitrary oriented rotation?}

\begin{figure}[!htb]

	\centering
\resizebox{0.9\linewidth}{!}{\input{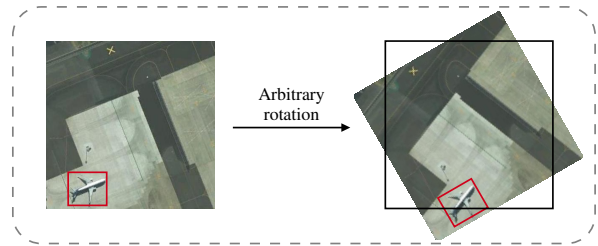}}
	\caption{In this case, for the airplane located near the edges of the image, a portion of the area will be lost during arbitrary rotations, making them unsuitable for regular training.
	}
	\label{fig:arbitrary_rotation}
\end{figure}

Due to the limited training samples in FSOD, objects near the edges of the image can be lost when dealing with objects rotated arbitrarily (see Fig. \ref{fig:arbitrary_rotation}). Therefore, we did not include arbitrary rotations in the transformations. Other geometric transformations might enhance the model's performance. We will validate this in our future work.

\subsubsection{Sensitive to selection of novel samples}

The sensitivity of FSOD to the selected support samples is also crucial. For results in Tab.~\ref{tab:supportchoice}, we carried out 10 runs with different support samples for FSCE~(second best model) and our method (TINet). We observe from Tab.~\ref{tab:supportchoice} that TINet is slightly better than FSCE and the results are relatively stable w.r.t the choice of support samples.
\begin{table}[!htb]

\footnotesize
	\caption{Comparing results of different FSOD methods on NWPU VHR-10.v2 test set (nAP).}

	\centering
        \setlength{\tabcolsep}{1.8mm}{
	\begin{tabular}{c|c|c|c|c|c} 

            \toprule
		Method&Backbone&2&3&5&10\\
		\midrule

        FSCE \cite{fsce}&ResNet-50&54.5$\pm$0.8&56.9$\pm$0.9&62.1$\pm$1.2&69.2$\pm$0.9\\
        \midrule
        TINet(ours)&ResNet-50&54.1$\pm$1.0&56.2$\pm$1.0&63.3$\pm$0.9&70.3$\pm$1.1\\

		\bottomrule
	\end{tabular}}
    \label{tab:supportchoice}
    
\end{table}

\subsubsection{Comparison with other transformation invariant methods}
We finally compared our method with two representative transformation invariant methods. ReResNet \cite{redet} achieves invariance by extracting rotation-invariant features, while TIP introduces Cutout and Gaussian Noise into the input images and utilizes consistency loss for achieving invariance. From Tab. \ref{tab:til_exp}, we can observe that the performance of the Strong Baseline deteriorates significantly after incorporating ReResNet. This is because ReResNet lacks a sufficient number of samples for training in a few-shot setting, leading to convergence issues. Moreover, the inference speed is significantly slower when using ReResNet (4.4 FPS compared to 15.5 FPS). As TIP \cite{tip} did not publish their source code, we managed to replace our geometric transformation with Cutout and Gaussian Noise, as proposed in TIP. The results suggest that geometric transformation is significantly superior to Cutout and Gaussian Noise, especially in the low-shot regime.

\begin{table}[h]

    \renewcommand\arraystretch{1.0}
	\caption{Comparing with other transformation invariant methods on the DIOR test set in split1 (nAP).$^\ast$ represents what we reproduced, and may differ slightly from the original paper}
    \footnotesize
	\centering
        \setlength{\tabcolsep}{1.6mm}{
	\begin{tabular}{c|c|c|c|c|c|c} 

		\toprule
		Method&Backbone&5&10&20&30&FPS\\
		\midrule


        Strong Baseline & ReResNet\cite{redet}&12.1&15.3&18.7&21.0&4.4\\
        TIP$^\ast$\cite{tip}& ResNet&26.6&33.4&40.1&41.3&15.5\\
        TINet(ours)&ResNet&29.5&35.2&41.6&42.8&15.5\\
        \bottomrule
	\end{tabular}}
    \label{tab:til_exp}

\end{table}

\begin{figure}
	
	\flushleft
	\renewcommand{\tabcolsep}{4pt}
	\begin{tabular}{c}
		\includegraphics[width=1.0\linewidth]{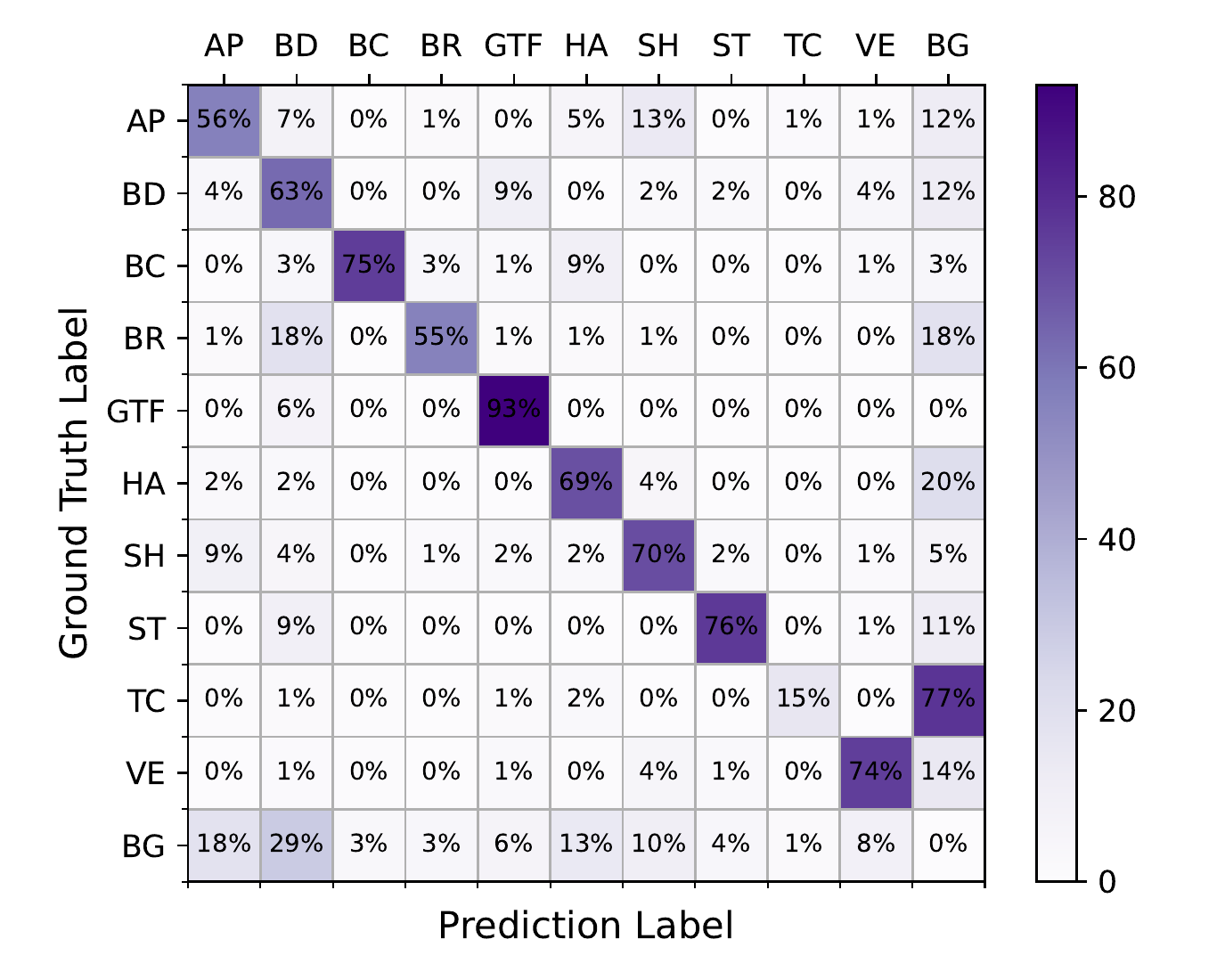}\\
		(a) \\
		\includegraphics[width=1.0\linewidth]{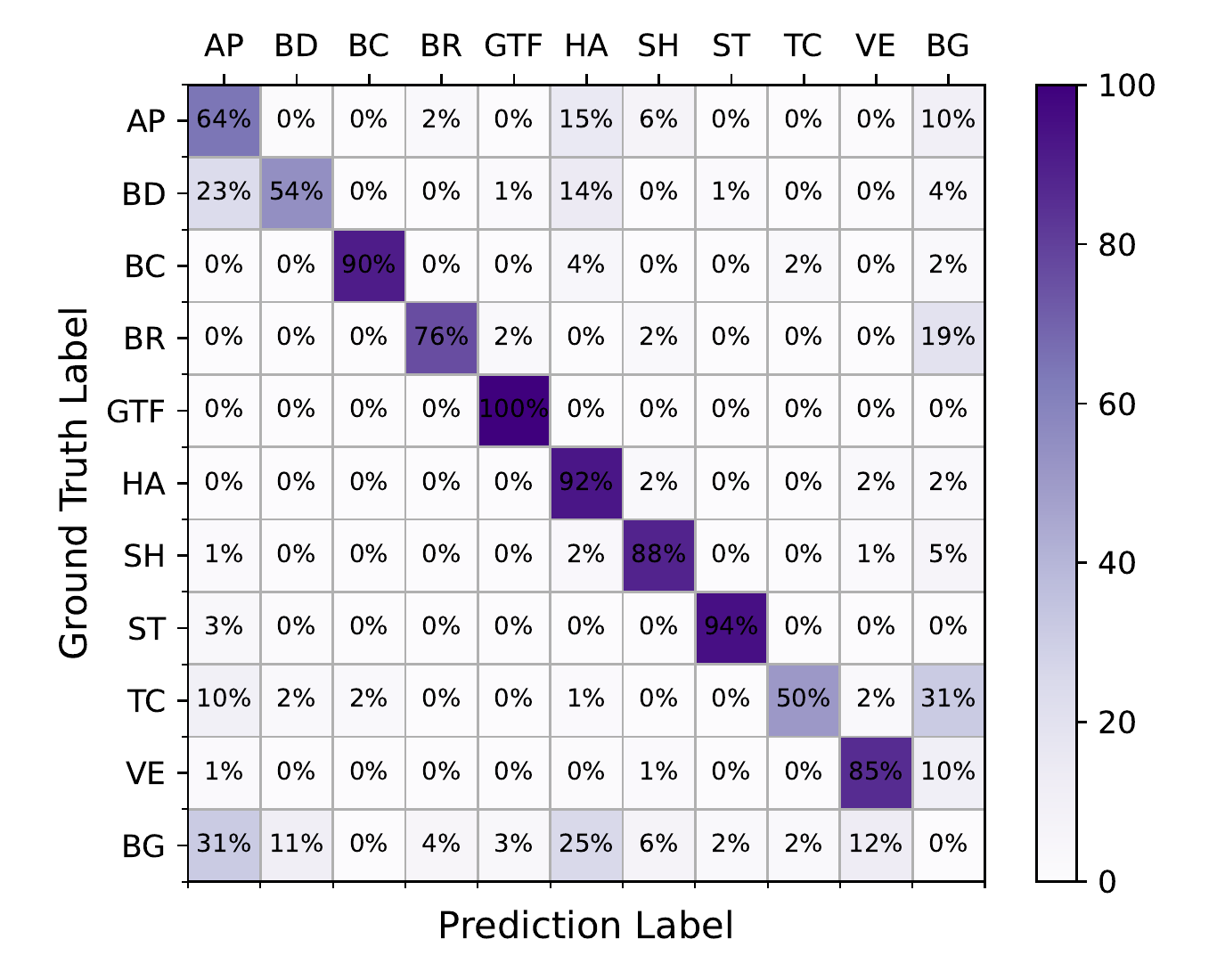}\\	
		(b) \\
	\end{tabular}
	
	\caption{Comparison of confusion matrix between the Strong Baseline and TINet. (a) Confusion matrix of the Strong Baseline. (b) Confusion matrix of TINet. The abbreviations for the categories are AP-airplane, BD-baseball diamond, BC-basketball court, BR-bridge, GTF-ground track field, HA-harbour, SH-ship, ST-storage tank, TC-tennis court, VE-vehicle, and BG-background.}
	\label{fig:cm_comparison}
\end{figure}

\begin{figure*}
	
	\centering

	\includegraphics[width=1.0\linewidth]{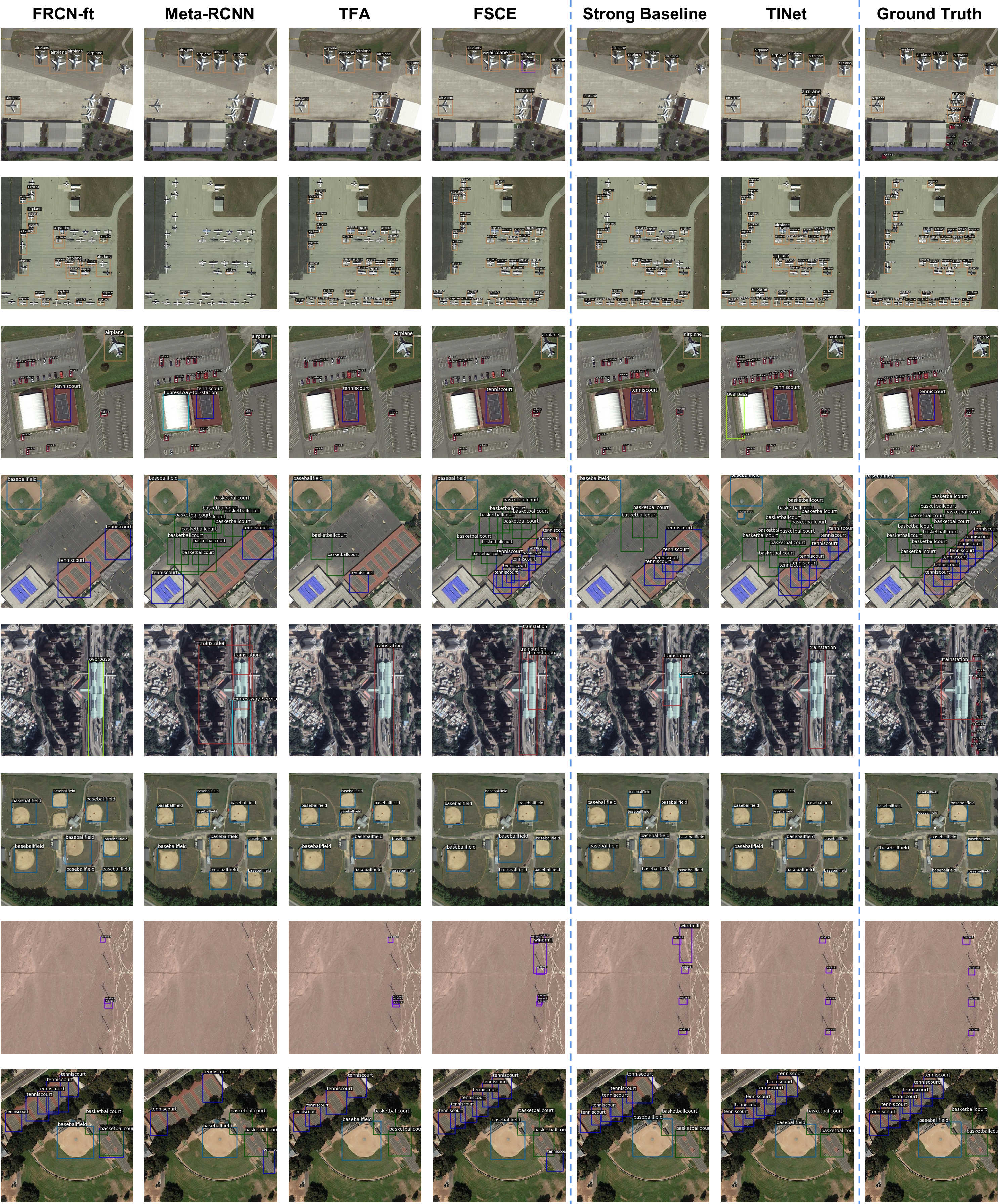}
	
	\caption{Detection results of different methods of sample images including the detected novel objects and a small number of base objects on DIOR dataset (30-shot at split1).}
    \label{fig:results_comparison_sup}
\end{figure*}

\begin{figure*}
	
	\centering

	\includegraphics[width=0.95\linewidth]{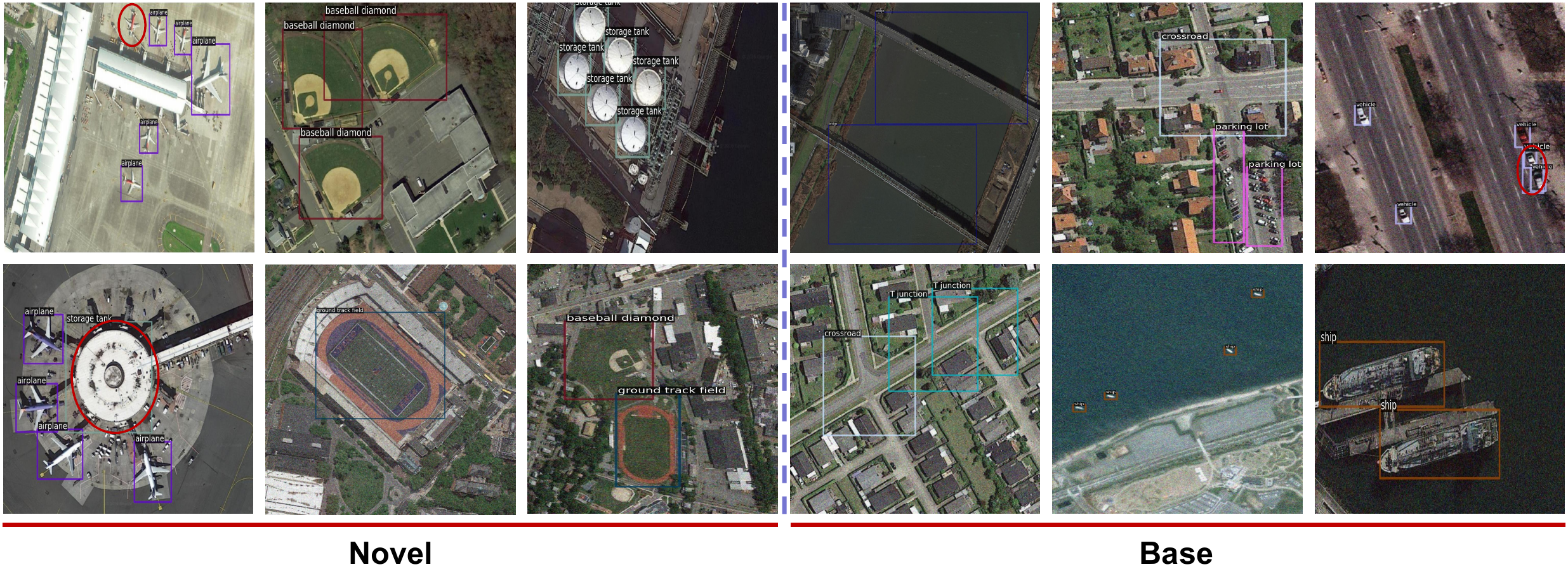}
	
	\caption{Detection results of the TINet for the novel and base objects in HRRSD dataset (30-shot). Objects of different categories are represented by rectangular boxes of different colors. Failure cases are represented by red ellipses.}
	\label{fig:vis_hrrsd}
\end{figure*}

\subsection{Visualization}
To more intuitively demonstrate the effectiveness of our method, we visualize the confusion matrix and prediction results.

\subsubsection{Confusion matrix}
We generated a confusion matrix using the detection results of the NWPU VHR-10.v2 test set. The abbreviations for the categories are as follows: AP-airplane, BD-baseball diamond, BC-basketball court, BR-bridge, GTF-ground track field, HA-harbour, SH-ship, ST-storage tank, TC-tennis court, VE-vehicle, and BG-background. Unlike the classification task, the detection task involves cases of false positives and missing detections, making the inclusion of a background class necessary to cover all cases. It should be noted that the percentage sum along the horizontal axis is 100\%, while the vertical axis is not 100\% due to normalization based on the horizontal axis. From Fig. \ref{fig:cm_comparison}, it can be observed that the Strong Baseline has a low probability of correctly recognizing the novel classes (AP, BD, and TC). Our method, on the other hand, alleviates this problem to some extent. Additionally, it is worth mentioning that our method does not forget the characteristics of the base class while training on the novel class.

\subsubsection{Prediction results}
 
As shown in \figref{fig:results_comparison_sup}, we present a comparison of several FSOD methods on the DIOR dataset of 30-shot at split1, which contains objects of the novel category, including airplanes, baseball fields, train stations, tennis courts, and windmills, as well as a small number of objects from the base category. The results highlight the strong generalization ability of our proposed method, TINet, attributed to its multi-scale feature structure and transformation invariant learning. Notably, we observe that for smaller objects like airplanes and windmills, Meta-RCNN without the FPN structure performs even worse than the original fine-tuning Faster-RCNN (FRCN-ft). However, the incorporation of the Strong Baseline significantly enhances the detection performance of Meta-RCNN, leading to similar results as FSCE. Moreover, leveraging the transformation invariant strategy atop the Strong Baseline, TINet further improves the detection performance for objects with varying orientations, such as airplanes and tennis courts. For simpler objects like baseball fields, which lack scale and orientation diversity, all the compared algorithms achieve comparable detection results. Overall, our TINet outperforms all competing methods, producing the best detection results. 

Furthermore, we provide additional qualitative results in Fig. \ref{fig:vis_hrrsd}, encompassing both novel and base objects on the HRRSD (30-shot) dataset. This dataset exhibits less complexity compared to DIOR, resulting in fewer false detections. For example, the appearance of the circular aircraft waiting hall is very similar to the storage tank, so it caused false detection. The color of the missing airplane is overlaid with more red, which makes it different from other airplanes. Hence, this phenomenon motivates us should focus on designing modules to extract more discriminative features in future work.

\section{Conclusion}
\label{sec:Conclusion}

In this paper, in light of the challenges in few-shot object detection~(FSOD) for remote sensing images~(RSIs), we first propose to modify from existing meta-learning-based FSOD method by incorporating FPN and depth-wise convolution. To improve the network's ability to align the feature of the support branch and query branch, we further propose to incorporate transformation invariance into the baseline, which is then referred to as TINet. Extensive experiments demonstrate the effectiveness of our method, and the method achieved state-of-the-art performances on the vast majority of the metrics on three widely used optical remote sensing object detection datasets, i.e., NWPU VHR-10.v2, DIOR, and HRRSD. It is worth noting that our work is to demonstrate that the improvement of the FSOD in RSIs by geometric transformation is significant. In general, more geometric transformations may further improve performance, such as arbitrary rotation, scaling, translation, etc, which will be considered in detail in our future work. Among them, arbitrary rotation transformation may introduce an artificial black border area and the risk of GT information leakage, which requires a special design.

\small
\bibliography{ref/ref}




\end{document}